# Transformer Network-based Reinforcement Learning Method for Power Distribution Network (PDN) Optimization of High Bandwidth Memory (HBM)

Hyunwook Park, Minsu Kim, Seongguk Kim, Keunwoo Kim, Haeyeon Kim, Taein Shin, Keeyoung Son, Boogyo Sim, *Student Member, IEEE*, Subin Kim, Seungtaek Jeong, *Member, IEEE*, Chulsoon Hwang, *Senior Member, IEEE* and Joungho Kim, *Fellow, IEEE*.

*Abstract*—In this article, for the first time, we propose a transformer network-based reinforcement learning (RL) method for power distribution network (PDN) optimization of high bandwidth memory (HBM). The proposed method can provide an optimal decoupling capacitor (decap) design to maximize the reduction of PDN self- and transfer impedance seen at multiple ports. An attention-based transformer network is implemented to directly parameterize decap optimization policy. The optimality performance is significantly improved since the attention mechanism has powerful expression to explore massive combinatorial space for decap assignments. Moreover, it can capture sequential relationships between the decap assignments. The computing time for optimization is dramatically reduced due to the reusable network on positions of probing ports and decap assignment candidates. This is because the transformer network has a context embedding process to capture meta-features including probing ports positions. In addition, the network is trained with randomly generated data sets. The computing time for training and data cost are critically decreased due to the scalability of the network. Thanks to its shared weight property and the context embedding process, the network can adapt to a larger scale of problems without additional training. For verification, the results are compared with conventional genetic algorithm (GA), random search (RS), and all the previous RL-based methods. As a result, the proposed method outperforms in all the following aspects: optimality performance, computing time, and data efficiency.

*Index Terms*—Decoupling capacitor, High bandwidth memory (HBM), Power distribution network (PDN), Reinforcement learning, Transformer network.

## I. INTRODUCTION

RECENTLY, artificial intelligence (AI) era requires over the terabyte-per-second (TB/s) bandwidth between the processing units and memories in the AI server [1], [2]. To meet the demands, a graphic processing unit (GPU)–high bandwidth memory (HBM) module has been regarded as a promising solution. The GPU-HBM provides over several TB/s bandwidth between GPU and HBM with 1024 I/Os [3], [4]. This TB/s scale bandwidth is possible due to the silicon interposer and through silicon via (TSV) technologies [5], [6]. Those enable fine pitch interconnections and high density in the I/O interface. However, due to the increasing data rate from 1 Gbps at gen 1 to 6.4 Gbps at gen 3 and the tremendous number of I/Os, switching power in the HBM I/O interface has been increased by generations [7], [8]. Moreover, the supply voltage level keeps shrinking to limit the power consumption [3]. Therefore, robust power distribution network (PDN) design becomes further challenging. It is essential to ensure signal quality by lowering power supply noise (PSN) fluctuation and power supply induced jitter (PSIJ) [9]–[11].

Decoupling capacitor (decap) design is one of the most important processes to suppress PSN and PSIJ by lowering PDN impedance [12]. Simultaneous switching noise (SSN) is the main noise source in the HBM I/O interface, which leads to severe eye distortion as shown in Fig. 1(a) [13], [14]. Large SSN can be generated when the simultaneous switching current drawn by I/O buffers meets the high peak anti-resonances of the VDDQ PDN. Therefore, it is necessary to optimize decap design to lower the PDN impedance in the broadband frequency range while minimizing the layout area to reduce the process cost as well as the power consumption [15]–[17].

However, decap design optimization is a combinatorial optimization problem that has high computational complexity. This is because the previous decap assignment affects the next assignment in terms of PDN impedance reduction [18]. A large scale of the PDN decap optimization requires extremely high computing cost. Because the size of PDN Z-matrix to be optimized is increased proportional to the square of the number of ports and proportional to the number of frequency points and data set size. In addition, the decap optimization requires high

This work was supported by National R&D Program through the National Research Foundation of Korea(NRF) funded by Ministry of Science and ICT(NRF-2020M3F3A2A01081587). This article is an expanded version from the IEEE Electrical Design of Advanced Packaging and Systems (EDAPS), Shenzhen, China, December 14–16, 2020 [DOI: 10.1109/EDAPS50281.2020.9312908]. *(Corresponding author: Hyunwook Park.)*

H. Park, M. Kim, S. Kim, K. Kim, T. Shin, H. Kim, K. Son, B. Sim and J. Kim are with the School of Electrical Engineering, Korea Advanced Institute of Science and Technology, Daejeon 34141, South Korea (e-mail: hyunwookpark@kaist.ac.kr; min-su@kaist.ac.kr; seonggukkim@kaist.ac.kr; keunwookim@kaist.ac.kr; taeinshin@kaist.ac.kr; haeyeonkim@kaist.ac.kr; keeyoung@kaist.ac.kr; boogyo@kaist.ac.kr; joungho@kaist.ac.kr).

S. Kim is with Samsung Electronics, Suwon, 16677, South Korea (email: subinn.kim@samsung.com)

S. Jeong and C. Hwang are with Missouri University of Science and Technology, Rolla, MO, 65409, USA (email: sjeong@mst.edu; hwangc@mst.edu).

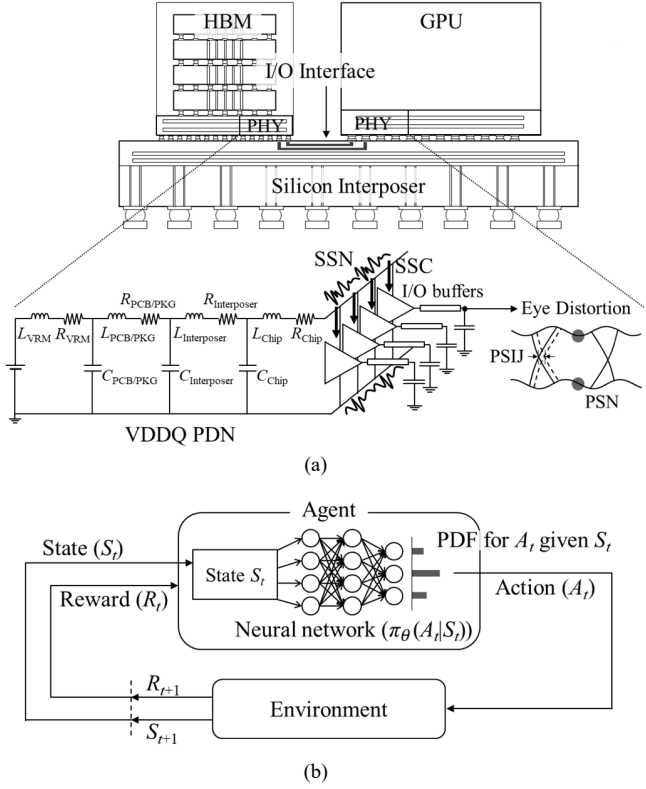

Fig. 1. (a) Eye distortion due to the large SSN generated by the numerous switching I/Os in the HBM-GPU I/O interface. (b) Reinforcement learning to solve combinatorial optimization problems defined by MDP parameters.

computing time since the computation of the PDN impedance is time-consuming. Also, massive iterations are needed for the optimization. In that point of view, the equivalent circuit modeling of the PDN should be preceded for the optimization, rather by the 3-D EM simulator [13]. Furthermore, a computing time and cost-effective methodology that is suitable for combinatorial optimization considering previous decap assignments is needed.

Various decap optimization methods have been investigated based on the conventional genetic algorithm (GA) [12], [16], [19]. However, the previous GA-based methods cannot capture sequential combinatorial relationships between the decap assignments which limits the optimality performance. Also, those cannot capture meta-features over problems such as the position of probing ports. The probing port is a power/ground (P/G) port to measure PDN impedances. Whenever a new problem is given, those must re-optimize it which requires massive iterations. Therefore, their lack of reusability causes an increment of the optimization time.

Recently, machine learning has been introduced for decap optimization [20]. Especially, RL-based methods are being actively studied [21]–[28]. As shown in Fig. 1(b), the RL is an algorithm to solve the problem defined by Markov decision process (MDP) parameters – state, action, reward, and policy [29]. The main goal of the RL is to learn the policy which determines the action for the given state, to maximize the reward by interaction with the environment. If the neural network is used to approximate scalar value functions and the policy is totally determined by the value function, it is called value-based RL and the network is a value network. If the network directly parameterizes the policy as probability distribution functions (PDFs), it is called policy-based RL and the network is a policy network. Deep Q learning-based on-board decap optimization methods using a multi-layer perceptron (MLP) as the value function approximator are investigated [23], [25]. Also, on-interposer and on-chip decap optimization using deep Q learning with a convolutional neural network (CNN) is studied [24]. However, those are limited in both the size of the solution space and the optimality performance. Because they are based on the value-based RL algorithms with simple value approximators such as the MLP and CNN. To increase the solution space size, direct policy parameterization by policy networks and training the networks through policy gradient algorithms are investigated [26], [27]. In addition, [27] introduced a transformer network to consider sequential relationships between the decap assignments for performance improvement. However, both of them do not have the reusability causing the increment of the optimization time. A recent study proposed a reusable method on the position of the probing port [28]. However, all the previous RL-based methods including [28] are limited in scalability which causes the increment of computing time and cost when training [21]–[28]. The scalability refers to the characteristic of whether the trained neural network model can respond to the scale of the problem. Moreover, all the previous RL-based methods only considered self-impedance for the design metric. Those exclude transfer impedance which indicates the coupled PSN [21]–[28]. In addition, multi-probing ports are not considered [21]–[25], [27], [28].

In this article, we propose a transformer network-based RL method for PDN decap optimization of the HBM. The proposed method can optimize decap design to maximize the reduction of both the self- and transfer impedance seen at the multiple ports. Consideration of the sequential relationships between the decap assignments using the transformer network achieves higher optimality performance. Moreover, the network has a context embedding process to capture meta-feature including positions of probing ports. The context embedding process increases the generalization ability to new problems. Therefore, the network trained with randomly generated data sets is reusable to new problems varying positions of probing ports and decap candidates. Due to the shared weight property and the context embedding, the network has the generalization ability on the scale of decap optimization problems. The reusability and scalability provide a significant reduction of the computing time and cost for both the training and optimization. By comparing with the conventional GA and RS methods and all the previous RL-based methods in [21]–[28], we validate the higher optimality and the lower computing cost and time.

II. PROPOSAL OF TRANSFORMER NETWORK-BASED RL METHOD FOR PDN DECOUPLING CAPACITOR OPTIMIZATION

In this section, a transformer network-based RL method for the PDN decap optimization is proposed. A brief explanation of the decap optimization problem for this work is as follows. It is denoted as a decap $n/m$ problem. The problem formulation of the decap $n/m$ is as follows: assigning $m$ number of unit NMOS decaps for given $n$ number of positions for decap assignment candidates $\mathbf{X}$ to maximize the reduction of the self- and transfer impedances seen at 4 probing ports $\mathbf{P}$. $\mathbf{X}$ is a set of feature vectors of $n$ decap ports from on-chip and on-interposer HBM VDDQ PDNs:

$$\mathbf{X} = \{\mathbf{x}_1, \mathbf{x}_2, \ldots, \mathbf{x}_n\}. \quad (1)$$

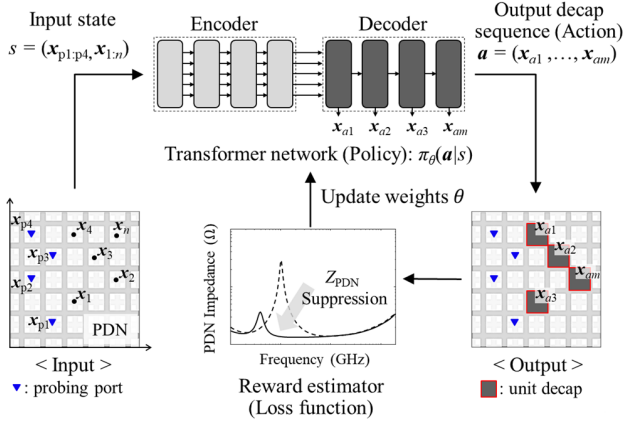

Fig. 2. Proposal of transformer network-based reinforcement learning method for PDN decap optimization.

$\mathbf{P}$ is a set of 4 probing ports in the I/O region of a physical layer (PHY) of the on-chip VDDQ PDN:

$$\mathbf{P} = \{\mathbf{x}_{p1}, \mathbf{x}_{p2}, \mathbf{x}_{p3}, \mathbf{x}_{p4}\}. \quad (2)$$

Therefore, a total 10 number of the self- and transfer impedances become a decap design criterion: $Z_{11}$, $Z_{22}$, $Z_{33}$, $Z_{44}$, $Z_{12}$, $Z_{13}$, $Z_{14}$, $Z_{23}$, $Z_{24}$ and $Z_{34}$. The total SSN is the sum of the self-noise and transfer-noises [5]. The amount of the self-noises is related to the self-impedances and that of the transfer-noises is related to the transfer impedances. Thus, all the 10 impedances need to be considered for the decap design criterion.

Fig. 2 shows the overall concept of the proposed transformer network-based RL method for the PDN decap optimization. For a given input state $s$, a policy network outputs a decap assignment sequence $\mathbf{a}$. The MDP parameters are defined as follows: State $s$ is a set of feature vectors of the probing ports $\mathbf{x}_{p1:p4}$ and the decap assignment candidates $\mathbf{x}_{1:n}$; Action $\mathbf{a}$ is deriving a decap assignment sequence $\mathbf{x}_{a1:am}$; Reward $r$ is modeled based on the reduction of the self- and transfer impedances compared to the initial PDN. The policy network is an encoder-decoder transformer network that is based on an attention mechanism [30]. The network directly parameterizes the probability distributions $\pi_\theta(\mathbf{a}|s)$ of $\mathbf{a}$ for the given $s$ by the weights $\theta$. Details on the definition of MDP parameters and the attention-based transformer network are explained in Section II-A and II-B respectively. The loop in Fig. 2 – the input $s$ goes into the transformer network; the network outputs $\mathbf{a}$; estimating the loss function of the PDN with assigned unit decaps by the reward estimator; updating the weight $\theta$ by the policy gradient algorithm to minimize the loss function – is iterated to optimize the policy network. Details on the training process of the network are described in Section II-C.

The network has the context embedding process to capture the meta-feature of the positions of the probing ports $\mathbf{P}$. The context embedding process increases not only the optimality, but also the generalization ability to new problems varying locations of $\mathbf{P}$ and the decap candidates $\mathbf{X}$. Details on the context embedding are explained in Section II-B. Moreover, the policy network is trained with randomly generated data sets for reusability. The data sets are the states and corresponding PDN Z-matrices, with randomly generated locations for $\mathbf{P}$ and $\mathbf{X}$. Details on the randomly generated data are described in Section II-D. The computing time for the RL-based methods is composed of training time to train the network and optimization time to obtain the optimized solutions. The reusability can reduce the optimization time because only one inference and one reward estimation are required without additional training for a new input.

The network has scalability on both the number of the decap candidates $n$ and that of the decap assignments $m$. $n$ is also called the size of the action space. The scalability is due to the context embedding and shared weight properties of the encoder and decoder networks. Details on the shared weight property are mentioned in Section II-B. The scalability can reduce the training time. Because the network trained in the smaller scale of the decap $n/m$ problems can be used to solve the larger scale of problems with the larger $n$ and $m$. Most of the training time is consumed in the impedance calculation by cascading the Z-matrices of PDN and decaps. Therefore, the size of $n$ and $m$ directly affects the training time.

*A. Markov Decision Process (MDP) for PDN Decap Design*

Fig. 3 shows the detailed explanation of the MDP parameters – state $s$, action $\mathbf{a}$ and reward $r$. Fig. 3(a) is the top view of the HBM VDDQ PDN. The PDN environment is configured of a 3-dimensional grid world, which is divided by 375×375 $\mu m^2$ unit decap-sized unit cells (UDUCs).

$s$ is defined as a set of 4-dimensional vectors $\mathbf{x}$'s of $\mathbf{P}$ and $\mathbf{X}$. Each $\mathbf{x}$ contains x-, y-, z-coordinates and whether the port is the probing port or not:

$$s = \{\mathbf{P}, \mathbf{X}\} = \{\mathbf{x}_{p1:p4}, \mathbf{x}_{1:n}\}$$
$$\mathbf{x}_i = (x\text{-coord}, y\text{-coord}, z\text{-coord}, isProbingPort). \quad (3)$$

where $i \in \{p1:p4, 1:n\}$ indicates a node index. $\mathbf{x}_{p1:p4}$ are denoted in blue triangles and $\mathbf{x}_{1:n}$ are in black circles in Fig. 3(a).

$\mathbf{a}$ is defined as deriving out the $m$ size of the decap assignment sequence as depicted in red circles in Fig. 3(a). Thus, it can be expressed as the permutation of sub-actions $a_t$:

$$\mathbf{a} = \{a_{t=1}, a_{t=2}, \ldots, a_{t=m}\} = \{\mathbf{x}_{a1}, \mathbf{x}_{a2}, \ldots, \mathbf{x}_{am}\}$$
$$\mathbf{x}_{ak} \in \mathbf{X}, \ k \in \{1:m\}. \quad (4)$$

The sub-action $a_t$ is defined as assigning a unit decap at a certain position $\mathbf{x}_{ak}$ at the time-step $t$. $\mathbf{x}_{ak}$'s are elements of $\mathbf{X}$. The unit decap is a 375×375 $\mu m^2$ sized NMOS decap with 1.055 nF capacitance and 0.7 mΩ equivalent series resistance (ESR). Details on the decap model are described in Section III.

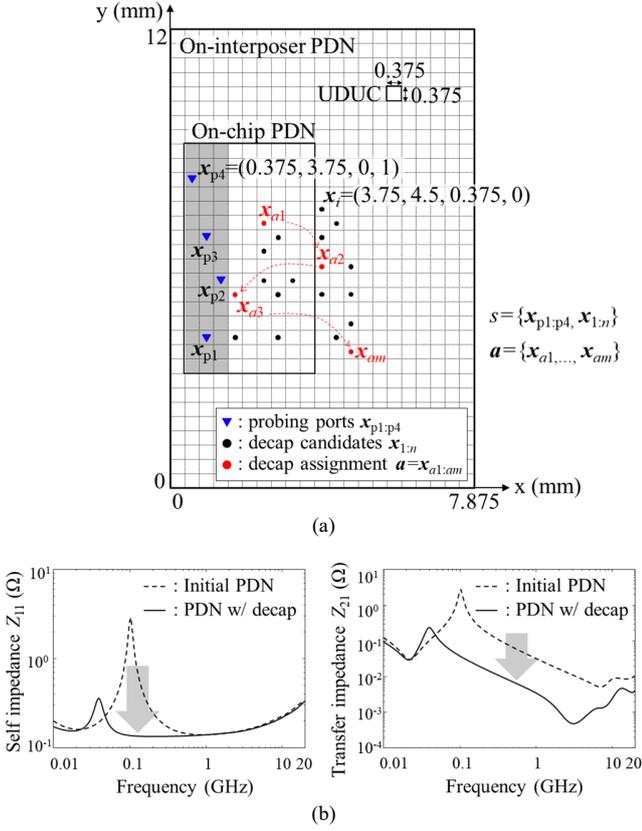

Fig. 3. Defined MDP parameters. (a) State *s* and Action *a*. (b) Reward *r*.

*r* is defined as a sum of the weighted mean of the self- and transfer impedance reduction at **P**:

$$r(\boldsymbol{a}|s) = \alpha_r \times \sum_{\mathbf{P}}\sum_{f}(|Z_{11\_initial}(f)|-|Z_{11\_wDecap}(f)|)/4 \\ +(1-\alpha_r)\times \sum_{\mathbf{P}}\sum_{f}(|Z_{21\_initial}(f)|-|Z_{21\_wDecap}(f)|)/6. \quad (5)$$

where $\alpha_r$ is the coefficient of the self-impedance reduction. $\alpha_r$ is set to 0.7 since usually the self-noise is more dominant than the transfer-noise [5]. *f* is a set of 166 frequency points in the logarithmic scale from 10 MHz to 20 GHz. $Z_{11\_initial}$ and $Z_{11\_wDecap}$ are the self-impedances seen at **P** of initial PDN and the PDN with decaps respectively; $Z_{21\_initial}$ and $Z_{21\_wDecap}$ are the transfer-impedances seen at **P** of the initial PDN and the PDN with decaps respectively, as shown in Fig. 3(b).

The policy $\pi_\theta(\boldsymbol{a}|s)$ is defined as a PDF of *a* for the given *s*. Since *a* is represented by the permutation of sub-actions $a_t$, it can be factorized by the PDFs of sub-actions $p_\theta(a_t|s, a_{1:t-1})$:

$$\pi_\theta(\boldsymbol{a}|s) = \prod_{t=1}^{m} p_\theta(a_t|s, a_{1:t-1}). \quad (6)$$

*B. Attention-based Encoder-decoder Transformer Network for Policy Approximation*

$\pi_\theta(\boldsymbol{a}|s)$ is approximated by the encoder-decoder transformer network as depicted in Fig. 4(a). The main mechanism of the network is the attention. By the attention computation, the encoder embeds the input *s* into high-dimensional node embeddings *h* and the decoder provides *a* by sequentially computing the PDF of the next assignment $p_\theta(a_t)$.

The attention mechanism is a weighted value passing algorithm between nodes to capture their relationships [31]. Every node can have query *q*, key *k* and value *v*. *q* is the object for questioning to the neighbor nodes and *k* is the description or characteristic for each node [30]. Therefore, each node can compute its compatibility *u* to the neighbor nodes by scaled dot-product of its *q* and *k* of the neighbor nodes [31, eq. (11)]. The attention weight *w* is computed by taking softmax function to *u* [30, eq. (1), 31, eq. (12)]. Finally, by using *w*, the next *h* can be embedded as the weighted sum of *v* of every node [31, eq. (13)]. Therefore, relationships between nodes can be captured in *h*. This computation is called a scaled dot-product attention or single head attention (SHA). A multi-head attention (MHA) is a parallel computation of scaled dot-product attention to learn diverse representation [30]. With *M* sets of linear projected *q*, *k* and *v*, each head is calculated through SHA. Finally, the concatenated *M* number of heads are linearly projected to the final node embedding *h*.

The main purpose of the encoder transformer network is to embed relationships between the 4 probing ports $x_{p1:p4}$ and the *n* decap candidates $x_{1:n}$ into *h*. It converts the 4-dimensional input *s* ($d_x$=4) into 128-dimensional *h* ($d_h$=128). Fig. 4(b) shows the details of the encoder transformer network. Initial node embedding $h_i^0$ is embedded by the linear projection of input node $x_i$:

$$h_i^0 = \mathbf{W}_x x_i + \boldsymbol{b}_x. \quad (7)$$

where $i \in \{p1:p4, 1:n\}$ indicates the node index; $\mathbf{W}_x \in \mathbb{R}^{d_h \times d_x}$ and $\boldsymbol{b}_x \in \mathbb{R}^{d_h}$ are learnable parameters. The node embeddings $h_i^l$ of the layer *l* are updated sequentially by attention layers. The encoder is configured of *L* number of the attention layers. Each of them consisted of one MHA sublayer and one feed-forward (FF) sublayer with residual connections [31]:

$$h_i'^l = h_i^l + \text{MHA}^l(h_{p1}^l, h_{p2}^l, h_{p3}^l, h_{p4}^l, h_1^l, h_2^l, ..., h_n^l). \quad (8a)$$

$$h_i^{l+1} = h_i'^l + \text{FF}^l(h_i'^l). \quad (8b)$$

where $l \in \{1:L\}$ indicates the layer index. $h_i'^l$ is the output of a MHA sublayer; $h_i^{l+1}$ is the output of a FF sublayer. The residual connection means $x+f(x)$ for preventing the gradient vanishing [32]. The FF sublayer has one hidden layer with the dimension $d_{\text{ff}}$, hence there are 4 learnable parameters $\mathbf{W}_{\text{ff},1}$, $\boldsymbol{b}_{\text{ff},1}$, $\mathbf{W}_{\text{ff},2}$, and $\boldsymbol{b}_{\text{ff},2}$. A ReLu activation is used for the FF sublayer. Especially in the MHA sublayer of the encoder, all the nodes have their own queries which are denoted in the red lines as shown in Fig. 4(b). This is for all the nodes to capture their relationships with each other through the attention.

Fig. 4(c) shows the details of the MHA sublayer. It shows how the node embedding of the first probing port $h_{p1}^l$ is updated by the MHA – graphical explanation of (8a). First of all, each node embedding generates its $q^{l,k}$, $k^{l,k}$ and $v^{l,k}$ by linear projection with learnable parameters $\mathbf{W}_Q^{l,k}$, $\mathbf{W}_K^{l,k}$ and $\mathbf{W}_V^{l,k}$:

$$q_i^{l,k} = \mathbf{W}_Q^{l,k} h_i^l, \quad k_i^{l,k} = \mathbf{W}_K^{l,k} h_i^l, \quad v_i^{l,k} = \mathbf{W}_V^{l,k} h_i^l. \quad (9)$$

where $k \in \{1, 2, \ldots, M\} \in \{1:M\}$ indicates a head index. $M$ is the number of heads. $d_k$ and $d_v$ are the dimensions of key and value. $\mathbf{W}_Q^{l,k} \in \mathbb{R}^{d_k \times d_h}$, $\mathbf{W}_K^{l,k} \in \mathbb{R}^{d_k \times d_h}$ and $\mathbf{W}_V^{l,k} \in \mathbb{R}^{d_v \times d_h}$ are the learnable parameters to make each head $h'^{l,k}_{p1}$, hence $\mathbf{W}_Q^l$, $\mathbf{W}_K^l \in \mathbb{R}^{Md_k \times d_h}$ and $\mathbf{W}_V^l \in \mathbb{R}^{Md_v \times d_h}$. After $q^l$, $k^l$ and $v^l$ are computed, attention weights $w_{p1,j}^{l,k}$ between the first probing node and other nodes are calculated in parallel:

$$w_{p1,j}^{l,k} = softmax\left(\frac{(q_{p1}^{l,k})^T k_j^{l,k}}{\sqrt{d_k}}\right). \quad (10)$$

where $j \in \{p1:p4, 1:n\}$ indicates the node index. Then, each head $h'^{l,k}_{p1}$ is computed by the weighted sum of the values by the attention weight:

$$h'^{l,k}_{p1} = \sum_j w_{p1,j}^{l,k} v_j^l. \quad (11)$$

where $j \in \{p1:p4, 1:n\}$ indicates the node index. Finally, all the heads are concatenated and linearly projected back into $d_h$ dimension by a learnable parameter $\mathbf{W}_O^l \in \mathbb{R}^{d_h \times Md_v}$. Also, the residual connection is added:

$$h'^l_{p1} = \mathbf{W}_O^l Concat(h'^{l,1}_{p1}, h'^{l,2}_{p1}, \ldots, h'^{l,M}_{p1}) + h^l_{p1}. \quad (12)$$

where $Concat$ is the concatenation operation.

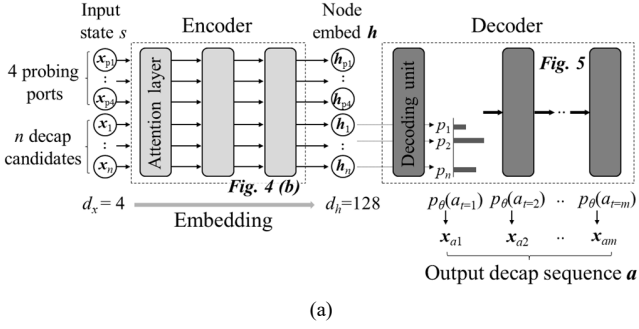

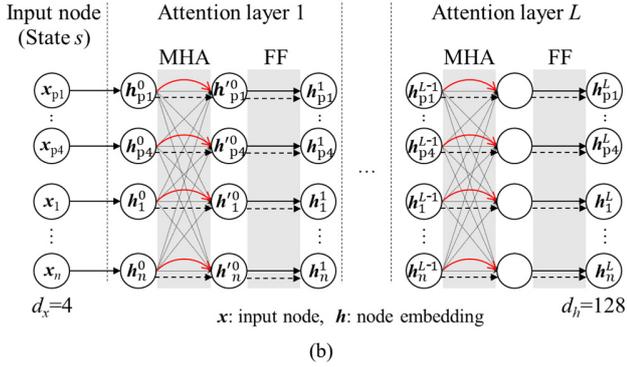

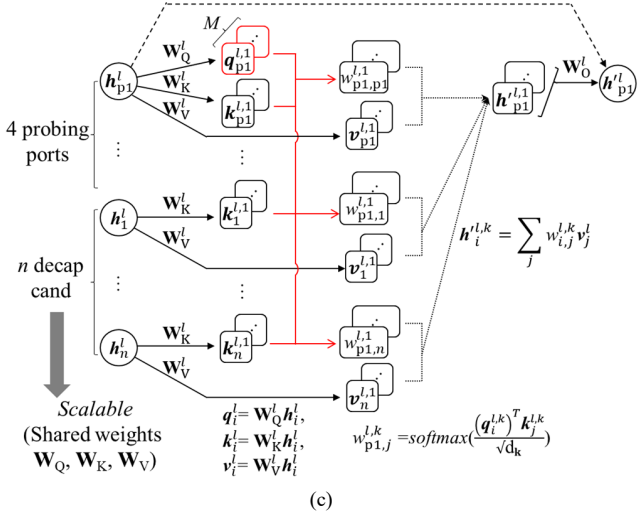

Fig. 4. (a) Encoder-decoder transformer network for policy approximation. (b) Encoder transformer network for embedding input state $s$ into high-dimensional node embeddings $h$. (c) Details of the attention layer of the encoder – MHA sublayer with residual connection.

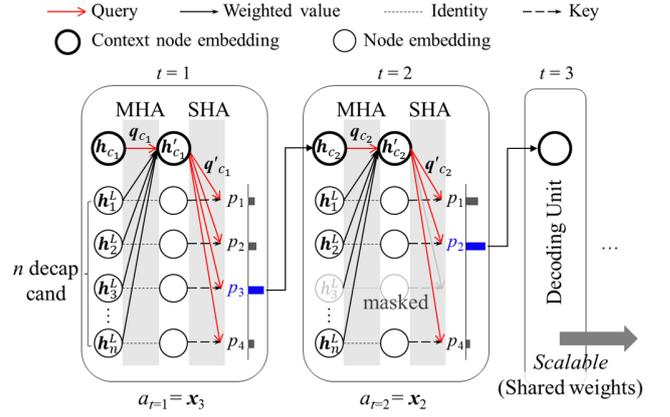

Fig. 5. Decoder transformer network providing the decap assignment sequence $a$ by computing the PDF of the next unit decap assignment.

All the learnable parameters $\mathbf{W}_x$, $b_x$, $\mathbf{W}_Q^l$, $\mathbf{W}_K^l$, $\mathbf{W}_V^l$, $\mathbf{W}_{ff,1}$, $b_{ff,1}$, $\mathbf{W}_{ff,2}$ and $b_{ff,2}$ are shared for each node in the same sublayer as depicted in Fig. 4(c). Therefore, the encoder transformer network is scalable on the number of the decap candidate nodes $n$. Because the nodes can be easily added without re-defining the learnable parameters.

Fig. 5 shows the details of the decoder transformer network. The decoder consists of the $m$ number of decoding units which is equal to the size of $a$. Each decoding unit computes $p_\theta(a_t)$ of the next decap assignment and outputs the assignment position $a_t$. A decoding unit is configured of one MHA layer and one SHA layer. Unlike the encoder, only context node embeddings $h_{c_t}$ and $h'_{c_t}$ become queries $q_{c_t}$ and $q'_{c_t}$. Then, $p_\theta(a_t|s, a_{1:t-1})$ is computed by the attention with keys $k$ from the encoder output node embeddings of the decap candidates $h_1^L$, $h_2^L$, …, $h_n^L$. Especially, the node embedding of the previous decap assignment $h_{a_{t-1}}^L$ is included in $h_{c_t}$ at every time-step as shown in Fig. 5. Therefore, the previous assignment can be considered to determine $a_t$. This is how the proposed method can capture the sequential combinatorial relationship between the decap assignments.

Since $h_{c_t}$ becomes the query and determines the policy for $a_t$, it is a key factor in the decoder. Thus, $h_{c_t}$ is defined as the following:

$$h_{c_t} = Concat(h_{a_{t-1}}^L, h_{p1}^L, h_{p2}^L, h_{p3}^L, h_{p4}^L, \bar{h}^L). \quad (13)$$

where $h_{a_{t-1}}^L$ indicates the node embedding of the previous decap assignment. $h_{p1}^L$, $h_{p2}^L$, $h_{p3}^L$ and $h_{p4}^L$ indicate node embeddings of 4 probing ports **P**. $\bar{h}^L$ is the mean of all the node embeddings. The dimension of the context node $d_{hc}$ equals to $d_h \times 6$. Therefore, not only the previous assignment, but also the positions of the probing ports can be considered when determining $a_t$.

Details on the derivation of $p_\theta(a_t|s, a_{1:t-1})$ are as follows. First, a new context node $h'_{c_t}$ is calculated in the MHA layer – $q_{c_t}$ projected from $h_{c_t}$; $k$ and $v$ from $h_1^L, h_2^L, \ldots, h_n^L$:

$$h'_{c_t} = MHA(h_{c_t}, h_1^L, \ldots, h_n^L)$$
$$q_{c_t} = W_Q h_{c_t}, \ k_i = W_K h_i^L, \ v_i = W_V h_i^L. \quad (14)$$

where $W_Q \in R^{Md_k \times d_{hc}}$, $W_K \in R^{Md_k \times d_h}$ and $W_V \in R^{Md_v \times d_h}$ are the learnable parameters. $i \in \{1:n\}$ indicates the node index of the decap candidates. Then, the compatibility $u$ is computed by scaled dot-product between $q'_{c_t}$ and $k' - q'_{c_t}$ projected from $h'_{c_t}$, and $k'$ is from $h_1^L, h_2^L, \ldots, h_n^L$:

$$q'_{c_t} = W'_Q h'_{c_t}, \ k'_i = W'_K h_i^L$$
$$u_i = \begin{cases} C \tanh\left(\dfrac{(q'_{c_t})^T k'_i}{\sqrt{d_{k'}}}\right), & \text{if } i \neq a_{t'} \ \forall t' < t \\ -\infty, & \text{otherwise.} \end{cases} \quad (15)$$

where $W'_Q$, $W'_K \in R^{d_k \times d_h}$ are the learnable parameters. $i \in \{1:n\}$ indicates the node index of the decap candidates. $d_{k'}$ is a dimension of the key in the SHA layer. $C$ is a tanh clipping [33]. Selecting the locations of the previous assignments is masked by making $u$ be negative infinity. Finally, $p_\theta(a_t|s, a_{1:t-1})$ is provided as the softmax of $u$:

$$p_\theta(a_t = x_i|s, a_{1:t-1}) = p_i = \frac{e^{u_i}}{\sum_{j=1}^n e^{u_j}}. \quad (16)$$

where $i \in \{1:n\}$ indicates the node index of the decap candidates.

Based on the estimated $p_\theta$, there are two decoding strategies: greedy and sampling. The greedy selection is to choose the position where the probability is maximized – selecting the best action. The sampling selection is to choose according to the estimated $p_\theta$. In the training phase, only the sampling is used to balance exploration and exploitation because the agent cannot distinguish which is the decent action in the beginning of the training [29]. In the optimization phase, both the greedy and sampling selection can be used. The greedy can reduce the computing cost and time since only one inference and reward evaluation are required. In the sampling selection, the computing time and cost are increased as much as a multiple of the sampling width. However, the optimality performance is higher than that of the greedy because the best solution is selected among the sampled solutions.

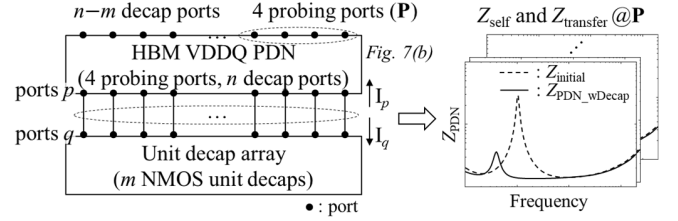

Fig. 6. Reward estimator to compute impedance of decap assigned PDN by cascading Z-matrix of HBM VDDQ PDN and that of unit decap array.

The decoder provides scalability on the number of the decap assignments $m$ as shown in Fig. 5. This is because all the learnable parameters $W_Q$, $W_K$, $W_V$, $W'_Q$ and $W'_K$ are shared for each decoding unit.

### C. Training Algorithm and Loss Function Calculation by the Reward Estimator

The policy gradient algorithm REINFORCE is used for training the policy network [34]. The loss function $\mathcal{L}(\theta|s)$ is the criterion for updating the weights $\theta$ of the network. Thus, the loss function $\mathcal{L}(\theta|s)$ is defined as the expectation of a cost function $L(a)$ which is defined as the negative reward:

$$\mathcal{L}(\theta|s) = \mathbb{E}_{\pi_\theta(a|s)}[L(a)] = \mathbb{E}_{\pi_\theta(a|s)}[-r(a|s)]. \quad (17)$$

As shown in Fig. 6, the reward $r(a|s)$ is calculated by the reward estimator which computes the impedance of the decap-assigned PDN $Z_{PDN\_wDecap}$. The reward estimator performs cascading the Z-matrix of the VDDQ PDN $Z_{PDN}$ and that of the decap array $Z_{da}$ consisting of $m$ unit decaps. The cascaded ports are the ports $p$ and $q$ in $Z_{PDN}$ and $Z_{da}$ respectively. Then, $Z_{PDN\_wDecap}$ is calculated using boundary conditions on the ports $p$ and $q$ [35, eqs. (1)–(3)].

Based on the defined $\mathcal{L}(\theta|s)$, the gradient of the loss function $\nabla_\theta \mathcal{L}(\theta|s)$ can be derived as the following equation [29], [34]:

$$\nabla_\theta \mathcal{L}(\theta|s) = \mathbb{E}_{\pi_\theta(a|s)}[(L(a) - L(a_{BL}))] \nabla_\theta \log \pi_\theta(a|s). \quad (18)$$

where $L(a_{BL})$ is the cost function estimated by the rollout baseline $\theta^{BL}$ [31]. The baseline network has the same configuration as the transformer policy network. It is periodically updated after every epoch if the improvement is critical according to a paired t-test [31]. The reason why subtracting the baseline is to reduce variance when training [29]. Finally, the learnable parameter $\theta$ of the policy network is optimized by a gradient descent Adam optimizer [36].

The policy network is trained with randomly generated data

**Algorithm 1** Training via REINFORCE with rollout baseline (decap $n/m$)

---

**Inputs:** action space size $n$, number of epochs $E$, steps per epoch $T$, batch size $B$, validation size $V$, full-ports PDN Z-matrix $\mathbf{Z}_{\text{fp}}$, $m$-sized decap array Z-matrix $\mathbf{Z}_{\text{da}}$

Initialize $\theta$, $\theta^{\text{BL}}$
**for** epoch = 1 to $E$ **do**
  **for** step = 1 to $T$ **do**
   $s_i, \mathbf{Z}_{si} \leftarrow \text{RandomDataGen}(n, \mathbf{Z}_{\text{fp}}) \ \forall i \in \{1, ..., B\}$
   $a_i \leftarrow \text{SampleInference}(s_i, \pi_\theta) \ \forall i \in \{1, ..., B\}$
   $a_i^{\text{BL}} \leftarrow \text{GreedyInference}(s_i, \pi_\theta^{\text{BL}}) \ \forall i \in \{1, ..., B\}$
   $r_i \leftarrow \text{RewardEst}(a_i, s_i, \mathbf{Z}_{si}, \mathbf{Z}_{\text{da}}) \ \forall i \in \{1, ..., B\}$
   $r_i^{\text{BL}} \leftarrow \text{RewardEst}(a_i^{\text{BL}}, s_i, \mathbf{Z}_{si}, \mathbf{Z}_{\text{da}}) \ \forall i \in \{1, ..., B\}$
   $\nabla_\theta \mathcal{L} \leftarrow \sum_{i=1}^B (L(a_i) - L(a_i^{\text{BL}})) \nabla_\theta \log(\pi_\theta(a_i))$
   $\theta \leftarrow \text{Adam}(\theta, \nabla_\theta \mathcal{L})$
  **end for**
  $s_j, \mathbf{Z}_{sj} \leftarrow \text{RandomDataGen}(n, \mathbf{Z}_{\text{fp}}) \ \forall j \in \{1, ..., V\}$
  $a_j \leftarrow \text{GreedyInference}(s_j, \pi_\theta) \ \forall j \in \{1, ..., V\}$
  $a_j^{\text{BL}} \leftarrow \text{GreedyInference}(s_j, \pi_\theta^{\text{BL}}) \ \forall j \in \{1, ..., V\}$
  **if** PairedTTest($\pi_\theta(a), \pi_\theta^{\text{BL}}(a^{\text{BL}})$) < 0.05 **then**
   $\theta^{\text{BL}} \leftarrow \theta$
  **end if**
**end for**

---

sets. This is to make reusable on the positions of **P** and **X**. Details on the random data set generator are explained in the following sub-section and an overall pseudo algorithm for training is shown in Algorithm 1.

*D. Random Data Set Generator for Reusability*

To train the reusable policy network, a random data set generator is implemented. It generates the data set $s$ and corresponding PDN Z-matrices $\mathbf{Z}_{\text{PDN}, s}$. $s$ is the set of state $s$ consisting of randomly selected 4 probing ports **P** and $n$ decap candidate ports **X** as shown in Fig. 3(a):

$$s = \{s_1, s_2, ..., s_N\}$$
$$s_k = \{\mathbf{P}_k, \mathbf{X}_k\}, \ k \in \{1, 2, ..., N\} \quad (19)$$
$$\mathbf{Z}_{\text{PDN}, s} = \{\mathbf{Z}_{s1}, \mathbf{Z}_{s2}, ..., \mathbf{Z}_{sN}\}.$$

where $N$ is the number of the data; $k$ indicates the data index; $\mathbf{P}_k$ and $\mathbf{X}_k$ are defined in (2) and (1) respectively. Each PDN matrix $\mathbf{Z}_{sk}$ is generated from a full-port Z-parameter model $\mathbf{Z}_{\text{fp}}$ of the HBM VDDQ PDN. $\mathbf{Z}_{\text{fp}}$ has total 816 ports on the on-chip and on-interposer PDN since one port is assigned for one UDUC in Fig. 3(a). 48 probing port candidates are in the I/O region (PHY) and 768 decap candidates are in the remaining on-chip and on-interposer PDN. $N$=1000, 100 and 100 are used for training, validation, and test sets respectively.

III. MODELING OF HBM VDDQ PDN FOR RANDOM DATA SET GENERATOR AND REWARD ESTIMATOR

For implementing the random data set generator and reward estimator, a VDDQ PDN model based on the HBM gen 2 and 2E is constructed as shown in Fig. 7(a) [37]–[39]. The modeling components considered in this work are as follows: on-chip grid

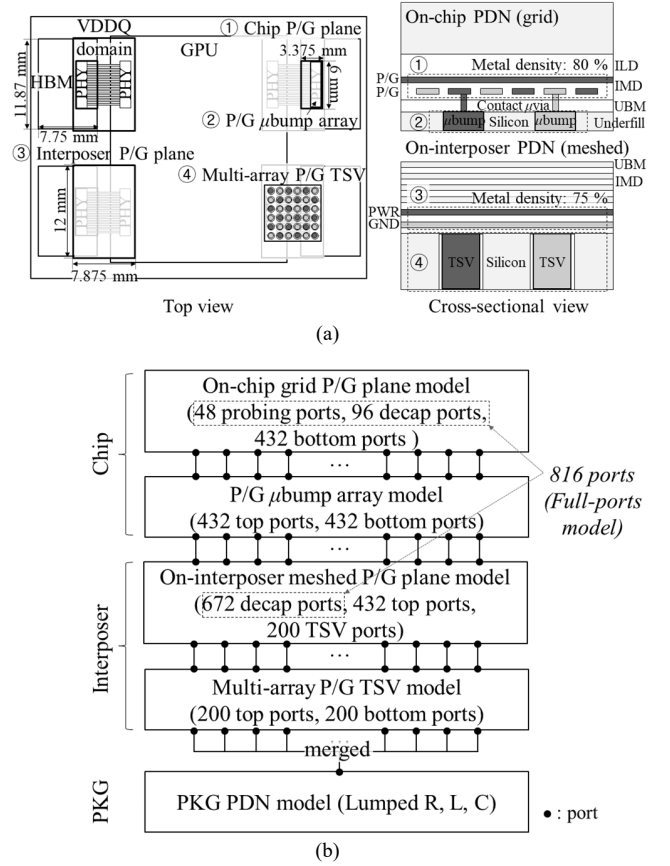

Fig. 7. (a) Configuration of HBM VDDQ PDN for this work. (b) Modeling of hierarchical VDDQ PDN impedance by cascading all the modeled components.

P/G planes, P/G $\mu$bump array with contact $\mu$vias, on-interposer meshed P/G planes, multi-array P/G TSVs, package (PKG) PDN, and NMOS decaps. Details on physical dimensions and material properties are summarized in Table I. Because the proposed method is to optimize the on-interposer and on-chip decaps, the precise and distributed models are implemented for on-chip and on-interposer PDNs. However, the lumped circuit model is enough for the PKG PDN to model the hierarchical PDN impedance profile. Also, it is more efficient in terms of computing cost when modeling for this work. The whole hierarchical PDN is modeled by cascading all the modeled components as shown in Fig. 7(b). The principle of cascading two Z-matrices is the same as described in Section II-C. All the distributed ports of the on-chip and on-interposer components are cascaded. Since PKG PDN is modeled as the lumped model, all the bottom ports of the multi-array TSV model are merged and cascaded to the PKG PDN [13].

The on-chip and on-interposer P/G planes are modeled by cascading the UDUCs. The UDUCs are modeled by cascading the UCs. Since the size of the UDUC is the same for both the chip and interposer, the on-chip UDUC is composed of 5625 (75 by 75) on-chip UCs and the on-interposer UDUC is configured of 625 (25 by 25) on-interposer UCs. Through W-element modeling methods, the UCs are modeled in per-UC resistance $R_{\text{UC}}$, inductance $L_{\text{UC}}$, conductance $G_{\text{UC}}$ and capacitance $C_{\text{UC}}$ as shown in Fig. 8(a) [40], [41]. Both $R_{\text{UC}}$ and $G_{UC}$ include DC and AC components ($R_0$, $R_f$, $G_0$, $G_f$). First, y-direction wave propagation of the UC is modeled. Using a perfect magnetic conductor boundary on the x-direction sides of the UC, a channel-like P/G plane with 2 ports on the other y-direction sides is made [40]. The channel-like P/G plane is simulated by 3-D EM simulator to obtain S-parameter. By assuming it as a W-element transmission line (TL) model, per-length model parameters $R_0$, $R_f$, $L$, $G_0$, $G_f$ and $C$ are extracted from the simulated S-parameter [41]. When extracting frequency-independent parameters $R_0$, $R_f$, $G_0$, and $G_f$ from real parts of per-length impedance $Z$ and admittance $Y$, least-square solutions are derived [41, eq. (6)]. When extracting frequency-dependent parameters $L$ and $C$ from imaginary parts of $Z$ and $Y$ respectively, the mean of the solutions is derived [41, eq. (9)]. The final UC model is obtained by extending the x-direction model parameters as same as the y-direction.

TABLE I
PHYSICAL DIMENSIONS AND MATERIAL PROPERTIES OF HBM VDDQ PDN

| Objective | Parameter | Description | Value |
|---|---|---|---|
| On-chip P/G plane | $L_{UC, chip}$ | Length of on-chip UC | 5 ($\mu m$) |
| | $W_{UC, chip}$ | Width of on-chip UC | 2 ($\mu m$) |
| | $S_{UC, chip}$ | Space of on-chip UC | 3 ($\mu m$) |
| | $t_{Cu, chip}$ | Metal (Copper) thickness | 0.5 ($\mu m$) |
| | $h_{Si, chip}$ | Height of silicon substrate | 30 ($\mu m$) |
| | $h_{IMD, chip}$ | Height of IMD layer | 0.5 ($\mu m$) |
| | $h_{ILD, chip}$ | Height of ILD layer | 2 ($\mu m$) |
| | $h_{UBM, chip}$ | Height of UBM layer | 1 ($\mu m$) |
| On-interposer P/G plane | $L_{UC, interposer}$ | Length of on-interposer UC | 15 ($\mu m$) |
| | $W_{UC, interposer}$ | Width of on-interposer UC | 7.5 ($\mu m$) |
| | $S_{UC, interposer}$ | Space of on-interposer UC | 7.5 ($\mu m$) |
| | $t_{Cu, interposer}$ | Metal (Copper) thickness | 1 ($\mu m$) |
| | $h_{Si, interposer}$ | Height of silicon substrate | 100 ($\mu m$) |
| | $h_{IMD, interposer}$ | Height of IMD layer | 1 ($\mu m$) |
| | $h_{UBM, interposer}$ | Height of UBM layer | 1 ($\mu m$) |
| $\mu$bump array | $d_{\mu bump}$ | Diameter of $\mu$bump | 25 ($\mu m$) |
| | $h_{\mu bump}$ | Height of $\mu$bump | 33 ($\mu m$) |
| | $p_{\mu bump}$ | Pwr-to-gnd pitch of $\mu$bump | 145.6 ($\mu m$) |
| Contact $\mu$via | $d_{\mu via}$ | Diameter of $\mu$via | 2 ($\mu m$) |
| | $h_{\mu via}$ | Height of $\mu$via | 1.5 ($\mu m$) |
| Multi-array P/G TSV | $d_{TSV}$ | Diameter of TSV | 15 ($\mu m$) |
| | $p_{TSV}$ | Pitch of TSV | 375 ($\mu m$) |
| | $t_{SiO2}$ | $SiO_2$ thickness | 0.5 ($\mu m$) |
| Material properties | $\sigma_{Cu}$ | Conductivity of Cu (P/G metal, $\mu$bump, contact $\mu$via) | $5.8 \times 10^7$ (S/m) |
| | $\sigma_{Si}$ | Conductivity of Si substrate | 10 (S/m) |
| | $\varepsilon_{IMD, chip}$ | Relative permittivity of on-chip IMD layer | 3.5 @ 9.0 GHz |
| | $\varepsilon_{ILD, chip}$ | Relative permittivity of on-chip ILD layer | 4.1 @ 9.0 GHz |
| | $\varepsilon_{UBM, chip}$ | Relative permittivity of on-chip UBM layer | 6.5 @ 9.4 GHz |
| | $\varepsilon_{underfill, chip}$ | Relative permittivity of on-chip underfill | 3.2 @ 9.4 GHz |
| | $\varepsilon_{Si}$ | Relative permittivity of Si | 11.9 @ 10 GHz |
| | $\varepsilon_{IMD, interposer}$ | Relative permittivity of on-interposer IMD layer | 4.1 @ 9.0 GHz |
| | $\varepsilon_{UBM, interposer}$ | Relative permittivity of on-interposer UBM layer | 6.5 @ 9.4 GHz |
| | $tan\delta_{IMD, chip}$ | Loss tangent of on-chip IMD layer | 0.001 @ 9.0 GHz |
| | $tan\delta_{ILD, chip}$ | Loss tangent of on-chip ILD layer | 0.001 @ 9.0 GHz |
| | $tan\delta_{UBM, chip}$ | Loss tangent of on-chip UBM layer | 0.001 @ 9.4 GHz |
| | $tan\delta_{underfill, chip}$ | Loss tangent of on-chip underfill | 0.02 @ 9.4 GHz |
| | $tan\delta_{IMD, interposer}$ | Loss tangent of on-interposer IMD layer | 0.001 @ 9.0 GHz |
| | $tan\delta_{UBM, interposer}$ | Loss tangent of on-interposer UBM layer | 0.001 @ 9.4 GHz |

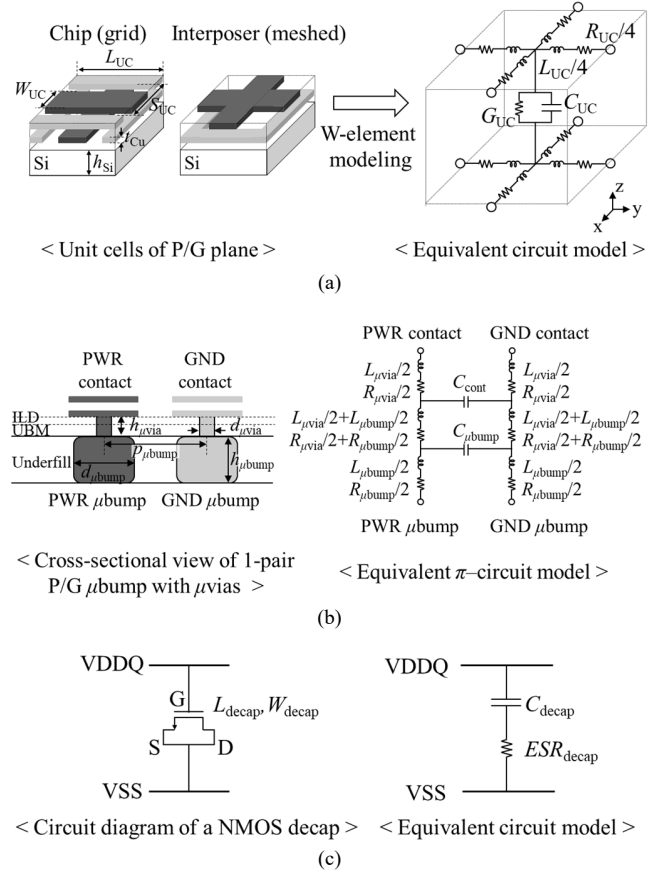

Fig. 8. (a) Modeling of on-chip and on-interposer UC. (b) Modeling of 1-pair P/G $\mu$bump with contact $\mu$vias. (c) Modeling of NMOS decap.

The $\mu$bump array in the HBM VDDQ PDN consists of 432 $\mu$bump pairs [37]. The coupling between $\mu$bump pairs is assumed to be negligible. Then, the model of $\mu$bump array $\mathbf{Z}_{\mu bumpArray}(f)$ is expressed by the extension of that of 1-pair $\mu$bumps $\mathbf{Z}_{1\text{-pair}}(f)$:

$$\mathbf{Z}_{\mu bumpArray}(f) = \begin{bmatrix} \mathbf{Z}_{1\text{-pair}}(f) & \cdots & \mathbf{0} \\ \vdots & \ddots & \vdots \\ \mathbf{0} & \cdots & \mathbf{Z}_{1\text{-pair}}(f) \end{bmatrix} \quad (20)$$

where all the diagonal elements are the 2-port $\mathbf{Z}_{1\text{-pair}}(f)$ and zero-matrices $\mathbf{0}$ otherwise. As shown in Fig. 8(b), the 1-pair $\mu$bumps with contact $\mu$vias are modeled as the equivalent $\pi$–circuit model. Both the $\mu$bump and $\mu$via pairs are modeled as cylindrical TLs. Detailed equations for model parameters $R_{\mu bump}$, $L_{\mu bump}$, $C_{\mu bump}$, $R_{\mu via}$, $L_{\mu via}$, and $C_{\mu via}$ are denoted in [42, eqs. (1)–(3)].

The multi-array TSV is configured of 400 P/G TSVs (20 by 20) with the case 2 pattern in [40, Fig. 10]. It is modeled based on the multi-conductor TL modeling method [40], [43]. Detailed equations are denoted in [40, eqs. (26)–(41)].

TABLE II
MODELED PARAMETERS OF THE HBM VDDQ PDN

| Objective | Parameter | Value |
|---|---|---|
| On-chip P/G plane | $R_{0,\text{chip}}/ R_{f,\text{chip}}$ | 30653.63/ 0.033 (Ω/m) |
| | $L_{\text{chip}}$ | 268.36 (nH/m) |
| | $G_{0,\text{chip}}/ G_{f,\text{chip}}$ | 0/ $1.18 \times 10^{-11}$ (S/m) |
| | $C_{\text{chip}}$ | 430 (pF/m) |
| On-interposer P/G plane | $R_{0,\text{interposer}}/ R_{f,\text{interposer}}$ | 2699.06/ 0.033 (Ω/m) |
| | $L_{\text{interposer}}$ | 120.74 (nH/m) |
| | $G_{0,\text{interposer}}/ G_{f,\text{interposer}}$ | 0/ $2.05 \times 10^{-11}$ (S/m) |
| | $C_{\text{interposer}}$ | 511.8 (pF/m) |
| μbump array | $R_{\mu\text{bump}}$ | 0.01 (Ω) @10 GHz |
| | $L_{\mu\text{bump}}$ | 8.1 (pH) |
| | $C_{\mu\text{bump}}$ | 1.2 (fF) |
| Contact μvia | $R_{\mu\text{via}}$ | 0.01 (Ω) @10 GHz |
| | $L_{\mu\text{via}}$ | 0.75 (pH) |
| | $C_{\mu\text{via}}$ | 0.03 (fF) |
| Unit NMOS decap | $C_{\text{unit\_decap}}$ | 1.055 (nF) |
| | $ESR_{\text{unit\_decap}}$ | 0.7 (mΩ) |
| PKG PDN (Lumped) | $R_{\text{PKG}}$ | 30 (mΩ) |
| | $L_{\text{PKG}}$ | 0.5 (nH) |
| | $C_{\text{PKG}}$ | 100 (nF) |

TABLE III
HYPER-PARAMETER SET-UP FOR THE TRANSFORMER-BASED
POLICY NETWORK AND TRAINING

| | Parameter | Value |
|---|---|---|
| Transformer-based Policy Network | Encoder | Attention layer # ($L$) | 3 |
| | | Input dim ($d_x$) | 4 |
| | | Embed (Hidden) dim ($d_h$) | 128 |
| | | Feed forward dim ($d_{ff}$) | 512 |
| | Decoder | Context node dim ($d_{hc}$) | 640 ($d_h \times 5$) |
| | | Hidden dim ($d_h$) | 128 |
| | | tanh clipping ($C$) | 10 |
| | Head dim ($M$) | 8 for MHA, 1 for SHA |
| | Key, value dim ($d_k$, $d_v$) | 16 for MHA, 128 for SHA |
| Training | Train epoch # ($E$) | 20 |
| | Train epoch size | 1000 |
| | Batch size ($B$) | 100 |
| | Validation size ($V$) | 100 |
| | Learning rate | $10^{-4}$ |
| | Learning rate decay | 0.95 |

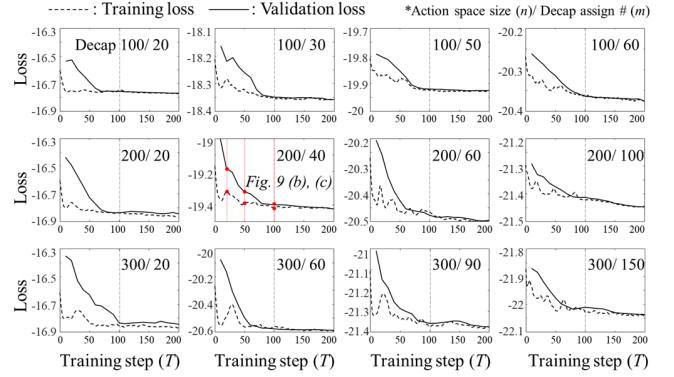

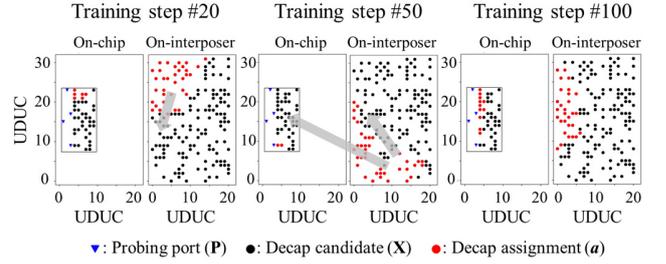

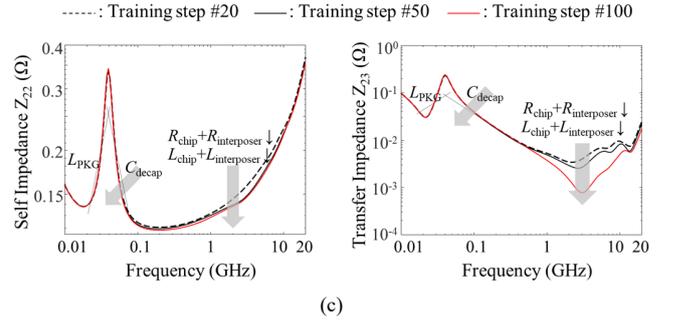

Fig. 9. (a) Training and validation loss convergence characteristics in 12 decap $n/m$ problems. (b) Decap optimization results depending on the training steps in test data #33 of decap 200/40. (c) Self- and transfer- impedance plots depending on the training steps in test data #33 of decap 200/40.

The unit NMOS decap for the sub-action $a_t$ is modeled by scaling the model of one NMOS decap. Generally, the NMOS decap is modeled as $C_{\text{decap}}$ and $ESR_{\text{decap}}$ in series as shown in Fig. 8(c) [42]. The model is based on the TSMC 65 nm process and the parameters $L_{\text{decap}}$ and $W_{\text{decap}}$ are 0.24 μm and 1 μm respectively. $C_{\text{decap}}$ and $ESR_{\text{decap}}$ for the decap are extracted by SPICE simulator. Then, $C_{\text{unit\_decap}}$ and $ESR_{\text{unit\_decap}}$ are scaled by the number of the decaps in the 375×375 μm² unit area. All the modeled parameters are summarized in Table II.

## IV. VERIFICATION OF THE PROPOSED METHOD

In this section, the superiorities of the proposed method is verified by comparing the optimality performance, reusability, scalability, computing time and cost to the previous methods including GA, RS and RL-based methods [21]–[28].

Table III summarizes the hyper-parameter set-up for the transformer network and training. The number of the attention layers in the encoder $L$ is set to 3. The dimensions of the input, embed (hidden), and feedforward layers in the encoder $d_x$, $d_h$, and $d_{ff}$ are set to 4, 128, 512 respectively. The dimensions of the context node and hidden layer $d_{hc}$ and $d_h$ in the decoder are set to 640 and 128 respectively. The value of tanh clipping $C$ is set to 10. The dimensions of the head, key and value $M$, $d_k$, and $d_v$ are set to 8, 16, and 16 for the MHA layers respectively and 1, 128, and 128 for the SHA layers respectively. The total number of 20 train epochs are trained; one epoch size is 1000; the batch size $B$ is 100 and the validation size $V$ is 100. The initial learning rate is $10^{-4}$ and is linearly decayed by 0.95 times every epoch. The proposed method is verified in 12 decap $n/m$ problems with the same hyper-parameter set-up: decap 100/20, 100/30, 100/50, 100/60, 200/20, 200/40, 200/60, 200/100,

TABLE IV
AVERAGE REWARD COMPARISON TO THE CONVENTIONAL
GENETIC ALGORITHM (GA) AND RANDOM SAMPLING (RS)

| Method<br>Problem | | Proposed Method | | GA | RS |
|---|---|---|---|---|---|
| Action space ($n$) | Decap num ($m$) | Sampling width | | | |
| | | 1 | 1000 | 1000 {100×10} | 1000 |
| 100 | 20 | 16.780 | *16.797 | 16.751 | 16.731 |
| | 30 | 18.364 | *18.370 | 18.339 | 18.326 |
| | 50 | 19.929 | *19.934 | 19.917 | 19.910 |
| | 60 | 20.381 | *20.386 | 20.361 | 20.358 |
| 200 | 20 | 16.849 | *16.888 | 16.808 | 16.765 |
| | 40 | 19.438 | *19.448 | 19.354 | 19.329 |
| | 60 | 20.509 | *20.515 | 20.443 | 20.422 |
| | 100 | 21.459 | *21.462 | 21.420 | 21.410 |
| 300 | 20 | 16.874 | *16.928 | 16.836 | 16.774 |
| | 60 | 20.589 | *20.604 | 20.473 | 20.440 |
| | 90 | 21.384 | *21.390 | 21.267 | 21.288 |
| | 150 | 22.055 | *22.057 | 21.979 | 21.991 |

*: Best optimality performance

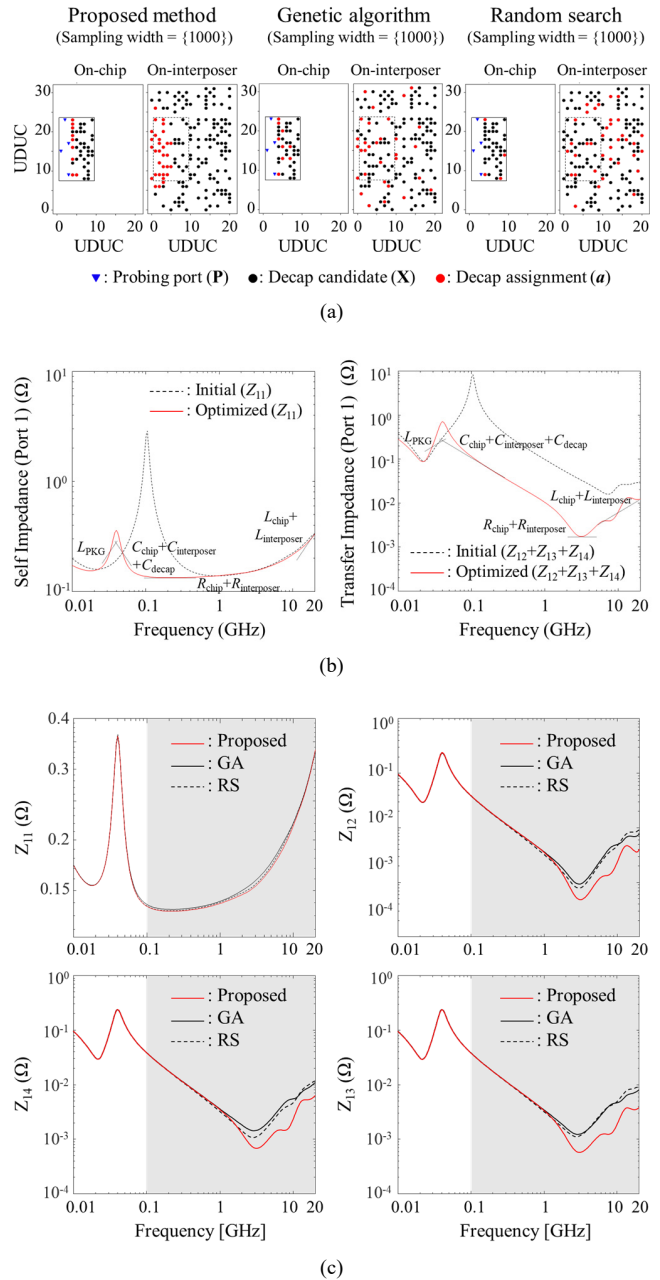

Fig. 10. (a) Comparison of decap optimization results of the test data #33 in the decap 200/40 between the proposed method, GA and RS. (b) Self and transfer impedances at the probing port 1 of the initial and the optimized PDN by the proposed method. (c) Comparison of the self and transfer impedances at the port 1 between the proposed method, GA and RS.

300/20, 300/60, 300/90 and 300/150.

*A. Training and Validation Loss Convergence Verification*

Fig. 9(a) shows the training and validation loss convergence characteristics in 12 decap $n/m$ problems. For every problem, training and validation loss converge after about 100 training steps (=100 batches=10 epochs=10 000 train data).

Fig. 9(b) and (c) show the performance improvement depending on the training steps in the decap 200/40– the decap optimization results and corresponding self- and transfer impedance of the test data #33. The blue triangles are 4 probing ports **P**, the black circles are the decap candidates **X** and the red circles are the decap assignment $a$ in Fig. 9(b). As the training progresses, the policy network assigns unit decaps near the probing ports in both the on-chip and on-interposer PDNs. This result coincides with the physical insight. Corresponding one self-impedance $Z_{22}$ and one transfer impedance $Z_{23}$ are plotted in Fig. 9(c). The probing ports are numbered starting from the lower to the upper port in the PHY region. As the training progresses, both the self- and the transfer impedance are reduced furthermore. From 40 MHz to 100 MHz frequency range, differences in the impedance profiles are not noticeable since the capacitance of the assigned decaps dominates. In other words, the number of the assigned unit decaps is the dominant factor. However, spreading loop resistance and inductance dominate in the frequency range over 100 MHz. Those are determined by the positions of assigned unit decaps. Therefore, the results of training step #100 where more unit decaps are assigned closer to the probing ports and between them show more reduced impedance profiles.

*B. Optimality Performance Verification of the Proposed Method by Comparison to the Conventional GA and RS*

In this sub-section, the optimality performance of the proposed method is compared to the conventional GA and RS [16], [28]. The average rewards of the 100 test data sets are compared in all the 12 decap $n/m$ problems. The 100 test data sets are generated by the random data set generator and are not correlated with the training and the validation sets. For a fair comparison, the reward functions of the GA and RS are set to be the same as the proposed method. Detailed algorithms of the GA and RS are explained in Algorithm 2 and 3 in the APPENDIX.

Table IV summarizes the average reward comparison results. For the same sampling width of 1000, the proposed method outperforms the GA and RS in all the 12 decap problems. For the GA, the number of the population $P$ and the generation $G$ are 100 and 10 respectively. Moreover, even with 1 sampling width, the average rewards of the proposed method are higher

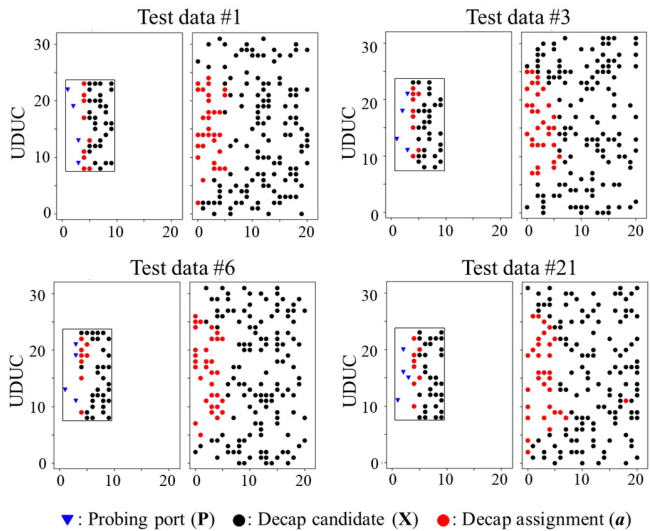

Fig. 11. Reusability verification on positions of the probing ports **P** and decap candidates **X** – decap optimization results of test data #1, #3, #6 and #21 in decap 200/40.

than those of the GA and RS with the sampling width of 1000 in all the problems.

Detailed results of the test data #33 in the decap 200/40 is shown in Fig. 10. The results when all the methods have the sampling width of 1000 are compared – same computing time for the optimization. Fig. 10(a) depicts the decap optimization results. The distribution of the assigned unit decaps is more confined near the probing ports for both the on-chip and on-interposer PDN in the proposed method than the GA and RS. Fig. 10(b) shows the self and transfer impedances at the probing port 1 of the initial and optimized PDN by the proposed method. The reduction of $Z_{11}$ indicates the decrement of the self-noise. The reduction of the summation of transfer impedances $Z_{12}+Z_{13}+Z_{14}$ indicates that of the transferred noises from port 2, 3 and 4. For both the self- and transfer impedances, the assigned decaps reduces the anti-resonance between $L_{PKG}$ and $C_{chip}+C_{interposer}$ around 100 MHz. Also, those reduces the resistance and inductance of the on-chip and on-interposer PDNs over 100 MHz. Fig. 10(c) depicts the comparison of the self- and transfer impedances at the port 1 between the proposed method, GA and RS. The red lines are the results of the proposed method. The black solid lines are those of the GA and the black dashed lines are those of the RS. For both the self- and transfer impedances, the proposed method suppresses impedance more than the GA and RS, especially in the frequency range from 100 MHz to 20 GHz where loop inductance and resistance are dominant.

*C. Reusability and Scalability Verification of the Proposed Method*

In this sub-section, the reusability and scalability of the proposed method are verified. The average rewards in Table IV are the results of the 100 randomly generated test sets that are not correlated to the training and validation sets. The results by the proposed method show superior performance than the GA and RS. Fig. 11 shows the detailed decap optimization results on 4 different test data #1, #3, #6 and #21 in the decap 200/40. Regardless of the positions of the probing ports **P** and decap candidates **X**, the reusable policy network provides well-designed results without additional training. Therefore, the transformer-based policy network is reusable on the positions of **P** and **X**.

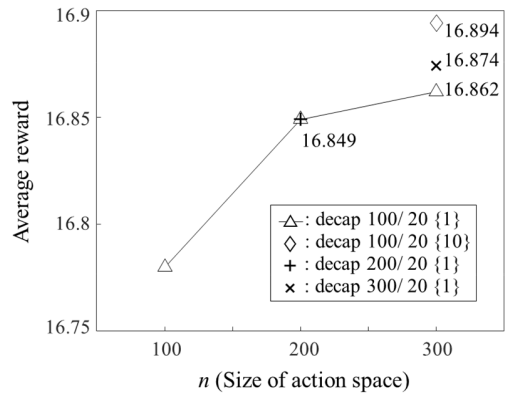

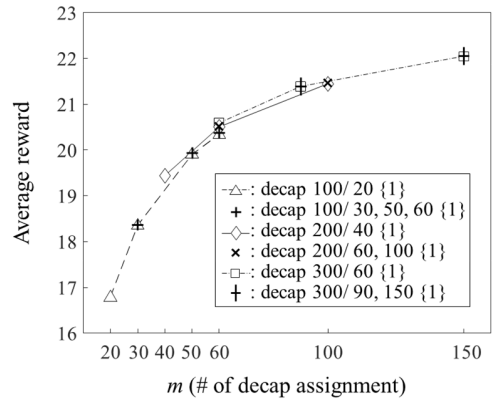

Fig. 12. (a) Verification of encoder scalability on the size of action space *n*. (b) Verification of decoder scalability on the number of the decap assignments *m*.

Fig. 12(a) shows the verification of the encoder scalability on the size of the action space *n*. Average rewards of the policy network trained in the smaller *n* and that trained in the original *n* are compared. The network trained in decap 100/20 is applied to solve the decap 200/20 and 300/20 problems. For the size of the action space of 200, the average rewards are equal to 16.849 without the sampling decoding strategy. For the size of the action space of 300, the average reward is slightly lower when the sample widths are the same as 1 – 16.862 for the network trained in decap 100/20; 16.874 for the network trained in decap 300/20. However, this gap is not noticeable and can be easily overcome by increasing the sampling width to 10. The average reward of the network trained in decap 100/20 with the sampling width 10 is 16.894. Thus, the network trained in decap 100/20 can adapt to solve the problems that have a larger *n*.

Fig. 12(b) depicts the verification of the decoder scalability on the number of the decap assignments *m*. The method is the same as the encoder scalability verification, but sweeping the size of *m* with the fixed *n*. The network trained in decap 100/20

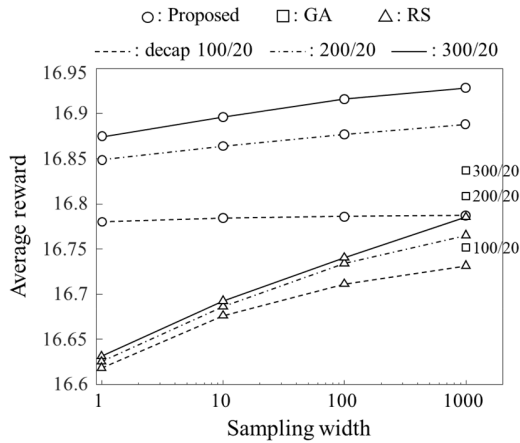

Fig. 13. Time-performance analysis on the proposed method and conventional GA and RS-based optimization methods.

TABLE V
COMPARISON OF OPTIMIZATION TIME AND COST FOR SAME PERFORMANCE BETWEEN THE PROPOSED METHOD, GA AND RS

| Method<br>Problem | | Proposed Method {1} | GA {1000 >} | RS {1000 >} |
|---|---|---|---|---|
| Decap 100/20 | Time | 11 s | 3 h > | 3 h > |
| | Cost | 2.8 GB | 2.8 TB > | 2.8 TB > |
| Decap 200/20 | Time | 57 s | 16 h > | 16 h > |
| | Cost | 10.8 GB | 10.8 TB > | 10.8 TB > |
| Decap 300/20 | Time | 2m 16s | 1d 14h > | 1d 14h > |
| | Cost | 24.6 GB | 24.6 TB > | 24.6 TB > |

* Total optimization time = 1t × sampling width {}, total cost= 1dc × sampling width {}
*1t= optimization time for 100 test data, 1dc= data cost of 100 test data
*CPU: Intel Xeon(R) Silver 4210R @2.4 GHz, GPU: NVIDIA GeForce RTX 3090, RAM: DDR4 512 GB

is applied to solve decap 100/30, 50, 60; that trained in decap 200/40 to solve decap 200/60, 100; that trained in decap 300/60 to solve decap 300/100, 150. It is to verify whether the network trained in problems assigning 20 % of the decap candidates is scalable to solve assigning up to 50–60 % of the decap candidates. The results in Fig. 12(b) show the same optimality performance in all the cases. Therefore, the decoder has the scalability on $m$.

### D. Computing Time and Cost Comparison to the Conventional GA, RS and Previous RL-based methods

In this sub-section, the computing time and cost of the proposed method are compared with the previous methods. The optimization time and cost are compared between the proposed method, GA and RS. The training time and cost are compared between the proposed method and the previous RL-based methods [21]–[28].

To compare with the GA and RS, a time-performance analysis is performed as shown in Fig. 13. The time-performance analysis is widely used to evaluate the performance improvement of the optimization methods depending on the time or sampling width [33]. The average rewards are compared to find out the values of the sampling width to realize the same performance. For every problem, even with the sampling width of 1000, the GA and RS do not reach the same performance of the proposed method with the sampling width of 1. In other words, the GA and RS need to sample 1000 times more to achieve the same performance. Therefore, more than 1000 times of the computing time and cost are required in the GA and RS. Detailed comparison results of the optimization time and cost to realize the same performance are summarized in Table V.

TABLE VI
COMPARISON BETWEEN THE PROPOSED METHOD AND PREVIOUS RL-BASED METHODS

| Method | Reusability | Scalability | RL type | NN | Solution space | Metric | Multi probe |
|---|---|---|---|---|---|---|---|
| Proposed | Yes | Yes | P | Transformer | $1 \times 10^{89}>$ | $Z_{11}/Z_{21}$ | Yes |
| [21] | No | No | V | - | $6.6 \times 10^{4}$ | $Z_{11}$ | No |
| [22] | No | No | V | - | $3.2 \times 10^{9}$ | $Z_{11}$ | No |
| [23] | No | No | V | MLP | $1 \times 10^{5}$ | $Z_{11}$ | No |
| [24] | No | No | V | CNN | $1.4 \times 10^{17}$ | $Z_{11}$ | No |
| [25] | No | No | V | MLP | $1 \times 10^{14}$ | $Z_{11}$ | No |
| [26] | No | No | P | MLP | $3.4 \times 10^{32}$ | $Z_{11}$ | Yes |
| [27] | No | No | P | Transformer | $1 \times 10^{60}$ | $Z_{11}$ | No |
| [28] | Yes | No | P | Ptr net | $8.2 \times 10^{18}$ | $Z_{11}$ | No |

*P: Policy, V: Value, NN: neural network for value/policy approximation, Ptr net: pointer network

TABLE VII
COMPARISON OF TRAINING TIME AND DATA COST DEPENDING ON THE ENCODER AND DECODER SCALABILITY

| | Action Space | Model (Trained in) | Train time (s) | Data cost (GB) |
|---|---|---|---|---|
| Encoder | 200 | 100/20 | 2200 | 30.8 |
| | | 200/20 | 11400 | 118.8 |
| | | Reduction | 80.7 % | 74.1 % |
| | 300 | 100/20 | 2200 | 30.8 |
| | | 300/20 | 27200 | 270.6 |
| | | Reduction | 91.9 % | 88.6 % |
| Decoder | 100 | 100/20 | 2200 | 30.8 |
| | | 100/60 | 3800 | 30.8 |
| | | Reduction | 42.1 % | 0 % |
| | 200 | 200/40 | 11400 | 118.8 |
| | | 200/100 | 25200 | 118.8 |
| | | Reduction | 54.8 % | 0 % |
| | 300 | 300/60 | 66400 | 270.6 |
| | | 300/150 | 104800 | 270.6 |
| | | Reduction | 36.6 % | 0 % |

*Data cost: size of train/validation data sets

For the previous RL-based methods in [21]–[28], we cannot directly compare the results on the decap $n/m$ problems. Because the previous works are limited in the solution space coverage and only considered the self-impedance in one probing port as shown in Table VI. In addition, the methods in [21]–[27] do not have reusability since the value or policy networks are trained in one PDN data. The networks always must be re-trained whenever a new different PDN data is given. Therefore, their optimization time includes the training time, causing a significant increment of the total computing time than the reusable proposed method.

Unlike the methods in [21]–[28], the proposed transformer network has the scalability. [21]–[28] must train the networks in the same scale of the decap problems that they intend to solve. However, the proposed transformer network can be trained in the smaller scale of the problems and solve the larger-scale problems. Table VII shows the reduction of the training time and data cost depending on the scalability of the encoder and decoder. The results are based on the scalability verification in Fig. 12(a) and (b) of Section IV-C. The data cost is the size of the 1000 training and 100 validation data.

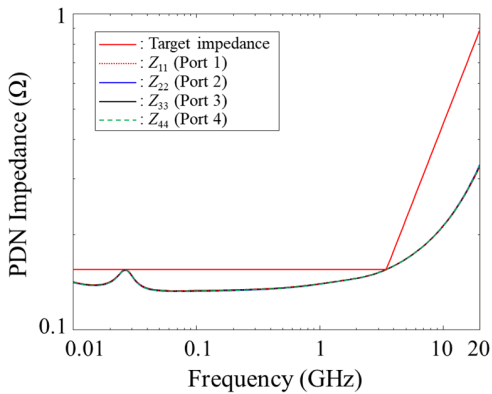

Fig. 14. Optimized self-impedance results by the proposed scalable transformer network to satisfy the R-L type target impedance.

The encoder scalability can reduce both the training time and data cost. Because the size of the VDDQ PDN Z-matrix for the training can be significantly reduced. It leads to the decrement of the execution time for cascading the VDDQ PDN and decap array. The reduction ratio of the training time between decap 100/20 and 200/20 is 80.7 %. That between decap 100/20 and 300/20 is 91.9 %. The reduction ratios of the data cost are 74.1 % and 88.6 % respectively. The decoder scalability can also reduce the training time because the size of the decap array Z-matrix is reduced. The reduction ratio of the training time between decap 100/20 and 60, that between decap 200/40 and 100 and that between decap 300/60 and 150 are 42.1 %, 54.8 % and 36.6 % respectively.

## V. Discussions

The proposed transformer network can be applied to satisfy the target impedance with the minimum layout area. The target impedance can be any type including widely used simple constant or R-L type [18]. Also, it can be derived from the PSIJ specification [10]. Rather training the network to be specialized to satisfy the certain target impedance as [21]–[27], the proposed network is trained to maximize the impedance reduction for the given number of decap assignments $m$. Also, the network is trained to be scalable on $m$. Therefore, the scalable network can provide minimized PDN impedance profiles for any given $m$. Then, by increasing the decap assignments, the network can find out the minimum number to meet the given target impedance.

Fig. 14 shows one example of the optimized self-impedance results by the proposed method to satisfy the R-L type target impedance. The scalable network trained in decap 300/150 is used to optimize HBM VDDQ PDN whose $n$ is 512. As shown in Fig. 14, all the self-impedances meet the given target impedance. The number of assigned unit decaps is 208 with 29.25 mm$^2$ of the layout area.

Regarding the generality on the size of the PDNs, the proposed network might have limitations on the coverage. Because the network is trained from one size of the PDN. It might still work on the different sizes of the PDNs since the network learned the heuristic to assign decaps referring to the probing ports. However, the coverage of the different sizes of PDN needs to be validated as further work. Moreover, to make the network more reusable and scalable on the size of the PDN, the network should be trained in various sizes of PDNs.

## VI. Conclusion

In this article, for the first time, we propose a transformer network-based RL method for the PDN decap optimization of HBM. The ability to capture sequential relations of the decap assignments, reusability and scalability realize superiorities on the optimality performance, computing time and cost, compared to the conventional GA, RS and the previous RL-based methods. In addition, the proposed method considers both the self- and transfer impedance seen at the multiple ports as optimization metrics. Therefore, we demonstrate the feasibility of the RL-based method in designing practical PDNs such as the HBM VDDQ PDN.

## Appendix

Detailed pseudo algorithms of the GA and RS for the decap $n/m$ problem are described in Algorithm 2 and 3 respectively. Detailed definitions of variables and functions in the GA are as follows:

1) **Variables in the GA**
   a) $P$: number of initial populations.
   b) $G$: number of generations.
   c) $s$: state of a test PDN.
   d) $Z_s$: Z-matrix of a test PDN.
   e) $Z_{da}$: Z-matrix of $m$-sized unit decap array.
   f) $\boldsymbol{ch^g}$ : a set of $P$ number of $g$-th generation chromosomes $\{ch_1^g, \ldots, ch_P^g\}$.
   g) $ch_i^g$ : an $i$-th chromosome in $\boldsymbol{ch^g}$ consisting of $n$ number of binary genes – 1: decap assigned, 0: no decap.
   h) $\boldsymbol{r^g}$: a set of rewards of $g$-th generation chromosomes $\{r_1^g, \ldots, r_P^g\}$.
   i) $r_i^g$: a reward of the $i$-th chromosome in $\boldsymbol{r^g}$.
   j) $\alpha$: ratio of selection for the next generation (=0.5 for this work).
   k) $\boldsymbol{sel^g}$ : a set of selected top $\alpha \times 100$ % of $g$-th generation chromosomes $\{sel_1^g, \ldots, sel_P^g\}$.
   l) $e$: ratio of elite chromosome selection for the next generation (=0.1 for this work).
   m) $\boldsymbol{elite^g}$ : a set of selected top $e \times 100$ % elite chromosomes which are remained without crossover.
   n) $\boldsymbol{lsel^g}$: a set of left halves of the selected chromosomes in $\boldsymbol{sel^g}$.
   o) $\boldsymbol{rsel^g}$ : a set of right halves of the selected chromosomes in $\boldsymbol{sel^g}$.
   p) $\beta$: probability to perform mutation (=0.05 for this work).
   q) $d$ : ratio of decaying the selection ratio $\alpha$ every generation (=0.95 for this work).

**Algorithm 2** GA for PDN decap optimization (decap $n/m$)

**Inputs:** action space size $n$, number of decap assignments $m$, number of initial population $P$, number of generations $G$, state of a test PDN $s$, Z-matrix of a test PDN $Z_s$, Z-matrix of $m$-sized unit decap array $Z_{da}$

Initialize $g = 0$
$ch^0 \leftarrow \text{RandomPopulation}(n, m, P)$
$r_i^0 \leftarrow \text{RewardEst}(ch_i^0, s, Z_s, Z_{da}) \; \forall i \in \{1, ..., P\}$
$ch^0 \leftarrow \text{Sort}(ch^0, r^0)$
$sel^0 \leftarrow \text{Selection}(\alpha, ch^0)$
$elite^0 \leftarrow \text{Selection}(e, ch^0)$
**for** $g = 0$ to $G - 1$ **do**
   $lsel^g \leftarrow \{sel_1^g[0:n/2], ..., sel_P^g[0:n/2]\}$
   $rsel^g \leftarrow \{sel_1^g[n/2+1:-1], ..., sel_P^g[n/2+1:-1]\}$
   $ch^{g+1} \leftarrow \{elite^g, \text{Crossover}(lsel^g, rsel^g)\}$
   **if** $rand() < \beta$ **then**
     $ch_i^{g+1} \leftarrow \text{Mutation}(ch_i^{g+1}) \; \forall i \in \{1, ..., P\}$
   **endif**
   $r_i^{g+1} \leftarrow \text{RewardEst}(ch_i^{g+1}, s, Z_s, Z_{da}) \; \forall i \in \{1, ..., P\}$
   $ch^{g+1} \leftarrow \text{Sort}(ch^{g+1}, r^{g+1})$
   $\alpha \leftarrow \alpha \times d$
   $sel^{g+1} \leftarrow \text{Selection}(\alpha, ch^{g+1})$
   $elite^{g+1} \leftarrow \text{Selection}(e, ch^{g+1})$
**end for**
**return** $argmax_{ch^{g+1}}(r^{g+1}), max(r^{g+1})$

---

**Algorithm 3** RS for PDN decap optimization (decap $n/m$)

**Inputs:** action space size $n$, number of decap assignments $m$, sampling width $N$, state of a test PDN $s$, Z-matrix of a test PDN $Z_s$, Z-matrix of $m$-sized unit decap array $Z_{da}$

$ch \leftarrow \text{RandomPopulation}(n, m, N)$
$r_i \leftarrow \text{RewardEst}(ch_i, s, Z_s, Z_{da}) \; \forall i \in \{1, ..., N\}$
**return** $argmax_{ch}(r), max(r)$

---

2) **Functions in the GA**
   a) RandomPopulation($n, m, P$): Random generation of $P$ number of $n$-sized chromosomes with $m$ number of 1-assigned genes.
   b) RewardEst($ch_i^g, s, Z_s, Z_{da}$): Reward estimator described in Section II-C.
   c) Sort($ch^g, r^g$): Sorting the chromosomes in $ch^g$ according to the reward $r^g$.
   d) Selection($\alpha, ch^g$): Selecting top $\alpha \times 100$ % of $g$-th generation chromosomes
   e) Crossover($lsel^g, rsel^g$): Generating (1- $e$) $\times P$ number of random combinations between $lsel^g$, $rsel^g$.
   f) Mutation($ch_i^g$): Exchanging two 0-assigned and 1-assigned genes.

The only different variable in the RS is a sampling width $N$ which indicates the number of population $P$ in the GA. Other variables and functions in the RS are the same as those explained in the GA.

ACKNOWLEDGMENT

We would like to acknowledge the technical support from ANSYS Korea.